\pdfoutput=1

\documentclass[11pt]{article}

\usepackage[final]{acl}

\usepackage[T1]{fontenc}

\usepackage[utf8]{inputenc}

\usepackage{microtype}

\usepackage{inconsolata}

\usepackage{graphicx}

\usepackage{times}
\usepackage{latexsym}

\usepackage{adjustbox}
\usepackage{multirow}
\usepackage{booktabs}
\usepackage{arydshln}

\usepackage{algorithm}
\usepackage{algorithmic}

\usepackage{enumitem}
\usepackage{amsmath,amssymb,amsfonts,bbm} 

\usepackage{subcaption}

\usepackage[capitalize]{cleveref}

\definecolor{commentcolor}{RGB}{110,154,155}   %

\setlist[itemize]{itemsep=0pt, parsep=0pt}

\definecolor{mygray}{RGB}{224,224,224}
\usepackage{colortbl}

\definecolor{lightblue}{RGB}{100,224,224}
\title{GAST: Gradient-aligned Sparse Tuning of Large Language Models with Data-layer Selection}

\author{
 \textbf{Kai Yao\textsuperscript{\rm 1}},
 \textbf{Zhenghan Song\textsuperscript{\rm 2}},
 \textbf{Kaixin Wu\textsuperscript{\rm 1}},
 \textbf{Mingjie Zhong\textsuperscript{\rm 1}},\\
 \textbf{Danzhao Cheng\textsuperscript{\rm 1}},
 \textbf{Zhaorui Tan\textsuperscript{\rm 3}},
 \textbf{Yixin Ji\textsuperscript{\rm 4}},
 \textbf{Penglei Gao\textsuperscript{\rm 5}\thanks{Corresponding authors.}},
\\
 \textsuperscript{1}Ant Group \ 
 \textsuperscript{2}Cornell University  \
 \textsuperscript{3}University of Liverpool \ \\
  \textsuperscript{4}Soochow University \
 \textsuperscript{5}Cleveland Clinic Lerner Research Institution \
\\
\href{mailto:jiumo.yk@antgroup.com}{jiumo.yk@antgroup.com}, \href{mailto:gaop@ccf.org}{gaop@ccf.org}
}

\begin{document}

\maketitle

\renewcommand{\thefootnote}{\fnsymbol{footnote}}
\begin{abstract}
Parameter-Efficient Fine-Tuning (PEFT) has become a key strategy for adapting large language models, with recent advances in sparse tuning reducing overhead by selectively updating key parameters or subsets of data. Existing approaches generally focus on two distinct paradigms: layer-selective methods aiming to fine-tune critical layers to minimize computational load, and data-selective methods aiming to select effective training subsets to boost training. However, current methods typically overlook the fact that different data points contribute varying degrees to distinct model layers, and they often discard potentially valuable information from data perceived as of low quality. To address these limitations, we propose Gradient-aligned Sparse Tuning (GAST), an innovative method that simultaneously performs selective fine-tuning at both data and layer dimensions as integral components of a unified optimization strategy. GAST specifically targets redundancy in information by employing a layer-sparse strategy that adaptively selects the most impactful data points for each layer, providing a more comprehensive and sophisticated solution than approaches restricted to a single dimension. Experiments demonstrate that GAST consistently outperforms baseline methods, establishing a promising direction for future research in PEFT strategies.
\end{abstract}

\section{Introduction}






\begin{figure}[!t]
\centering
\includegraphics[width=0.8\columnwidth]{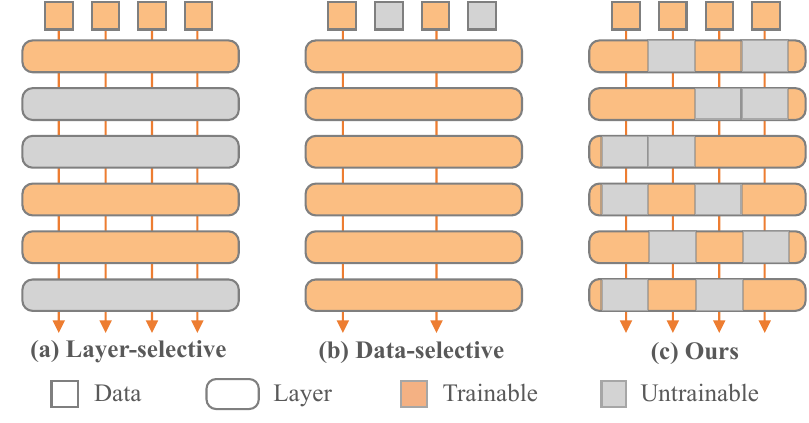}
\caption{Difference among (a) layer-selective methods, (b) data-selective methods, and (c) our method. layer-selective methods generate a subset of all layers to be updated with all mini-batch data. Data-selective methods utilize partial mini-batch data to train all layers. Our GAST selects a different subset of data for each layer.}
\label{fig:banner} 
\end{figure}

Large Language Models (LLMs) such as GPT-3~\cite{brown2020language} and LLaMA~\cite{touvron2023llama} form the backbone of modern NLP, yet their enormous size results in significant computational and memory overhead during full fine-tuning on downstream tasks~\cite{howard2018universal}.
To tackle this challenge, Parameter-Efficient Fine-Tuning (PEFT) methods have emerged as a mainstream solution, allowing users to adapt large models with significantly reduced resource overhead by tuning a subset of parameters \cite{series,lora,parallel,dora}. This PEFT strategy is largely guided by human-designed heuristics and fails to account for task-specific domain differences and nuances, which constrains its effectiveness across diverse downstream applications \cite{yao2024layer}. Despite these advantages, many PEFT methods still suffer from inefficiencies during training, often requiring substantial computation that undermines their intended performance benefits.

To further improve the efficiency of model training without compromising performance under the PEFT framework, numerous studies have been proposed to reduce unnecessary computation and enhance learning dynamics by integrating sparse tuning.
One prominent line of research focuses on layer selection, which posits that not all layers in an LLM are equally important for all training updates \cite{kaplun2023less,pan2024lisa,yao2024layer}. These approaches seek to reduce redundancy, such as optimizer memory, activation memory, and gradient memory, by estimating the importance score of each layer and selectively activating or updating only a subset of layers during training. While effective, most existing layer-wise methods apply a uniform layer configuration to all data samples within one mini-batch, implicitly assuming equal importance across samples and neglecting the inherent heterogeneity of data. Consequently, they may underutilize the representational capacity of the model for more complex or atypical samples.
Another influential direction centers on data selection, based on the observation that real-world datasets often contain large amounts of low-quality, redundant, or biased information \cite{wang2024greats,wang2024data}. By identifying and selecting a subset of informative data points, these methods aim to accelerate training and improve generalization. However, such approaches typically discard low-quality data entirely, potentially overlooking valuable information embedded in those seemingly uninformative examples—information that may become useful in later stages of learning or contribute to model robustness.

Although both layer selection and data selection offer promising routes for effective training and performance improvement, each suffers from critical limitations mentioned above when treated in isolation. 
One major empirical observation in this paper is that variations in data across tasks and domains might cause the phenomenon that different data points could make distinct and layer-specific contributions to model optimization. We hypothesize this is due to the fact that each layer of an LLM tends to capture different levels of semantic information. 
Therefore, using the whole dataset for all-layer fine-tuning may lead to gradient conflicts, leading to performance degradation. 
There remains a significant opportunity to develop more adaptive and fine-grained mechanisms that consider the interaction between data and model structure.

Motivated by the limitations of existing PEFT methods and the aforementioned observation, we investigate the data sparsity among the layers of LLMs and explore the interaction between data complexity and model depth.
Building on this insight, we present a theoretical framework that formally demonstrates that both layer selection and data selection are sub-optimal strategies to a more general joint selection paradigm. Thus, we propose a novel method, \textbf{G}radient-\textbf{a}ligned \textbf{S}parse \textbf{T}uning (GAST), which simultaneously performs layer-level and data-level selection. Fig.~\ref{fig:banner} shows the overall difference of our method compared to the layer-selective and data-selective methods. 
Specifically, GAST could dynamically sample a subset of data points that are most informative to that specific layer’s update and compute the gradient of the selected data points as the measurement. This enables us to preserve useful training signals from data points that might be discarded on other layers and ensures that each layer is trained on the most relevant and impactful samples. As a result, GAST achieves both layer-level sparsity and data-level importance, enhancing computational scalability while maintaining or even improving downstream task performance.

\begin{figure*}[!t]
\centering
\includegraphics[width=1.9\columnwidth]{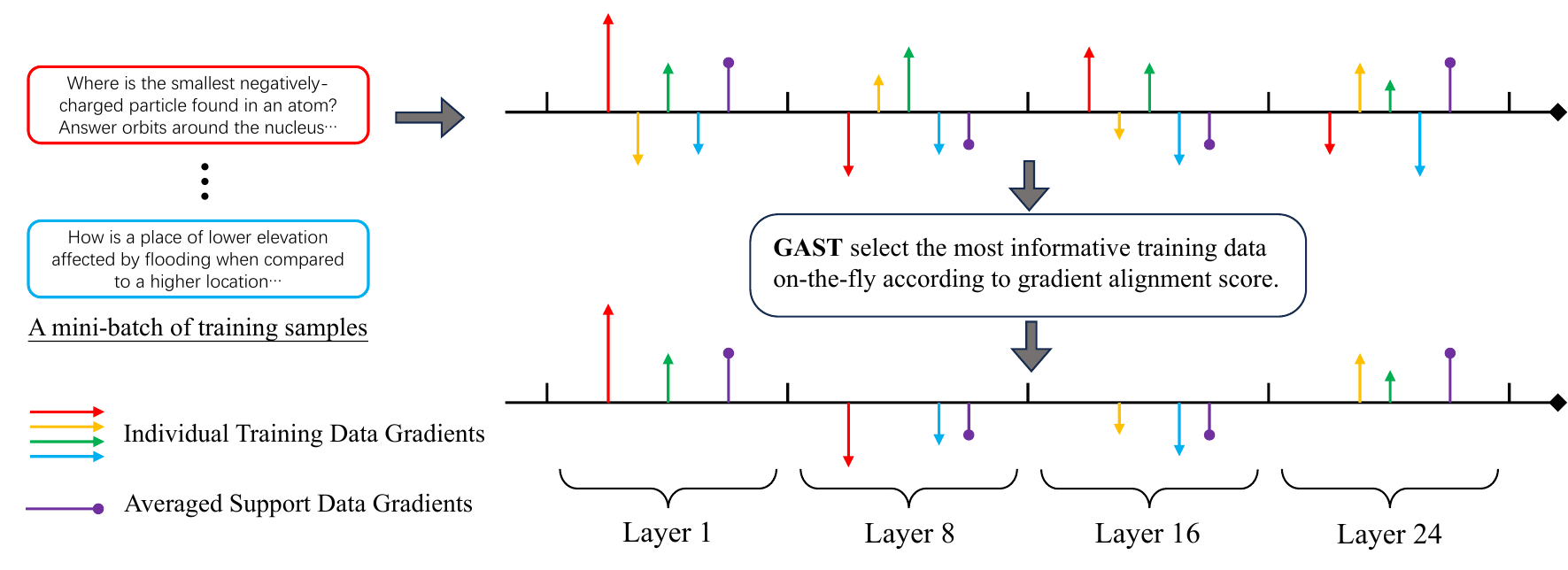}
\caption{Overall of our proposed Gradient-aligned Sparse Tuning (GAST). During training, every mini-batch exhibits gradient conflicts both among data samples and across model layers. To mitigate these conflicts, GAST uses the gradient of the support set to decide which individual sample should be used to update each layer. 
This data-layer selection reduces gradient interference and thereby improves both convergence speed and generalization performance.}
\label{fig:overall} 
\end{figure*}

In summary, our contributions are as follows:
\begin{itemize}
    \item We develop a theoretical foundation to demonstrate that both layer-level and data-level selection are sub-optimal to our hybrid data-layer selective sparse tuning.
    \item We propose a batch-level strategy that dynamically selects both data points and model layers to facilitate sparse training, effectively accelerating convergence and improving model performance.
    \item Extensive experiments across multiple LLMs and downstream tasks demonstrate that our proposed method achieves consistently better performance and faster convergence compared to existing PEFT approaches.
\end{itemize}


\section{Related Work}
\subsection{Parameter-efficient Fine-tuning}
PEFT has emerged as a dominant approach for adapting LLMs to downstream tasks while mitigating the prohibitive computational and memory costs of full fine-tuning. Early PEFT methods, such as adapter-based models \cite{series,parallel,lei2024conditional}, introduced small trainable bottleneck layers, allowing for task adaptation with minimal parameter updates. More recent methods like LoRA \cite{lora} and DoRA \cite{dora} further reduced the total number of trainable parameters. LoRA freezes the original model weights and injects low-rank trainable matrices into each layer’s weight update path, while DoRA further improves flexibility by decoupling the low-rank adaptation into separate scaling and shifting operations. Other techniques explore prompt-based tuning by tuning soft prompts while keeping the model frozen \cite{prefix,ptuning}. While PEFT methods have demonstrated strong empirical performance across diverse tasks, they typically assume uniform importance across training data and model layers, often failing to exploit the heterogeneity inherent in both. Our work builds on this foundation by introducing a dynamic mechanism that jointly considers layer-level and data-level sparsity, offering a more flexible strategy.

\subsection{Layer-wise Sparse Tuning}
Many studies have demonstrated that not all layers in an LLM contribute equally to task-specific adaptation and have proposed selectively fine-tuning a subset of layers to improve training efficiency \cite{fan2021layer,layerdrop,elhoushi2024layer}.
For instance, \citet{kaplun2023less} employs a greedy search strategy to identify the optimal layers for fine-tuning, but this process incurs substantial computational cost and initialization time. In the work of \cite{pan2024lisa}, the authors develop a new optimization strategy to accelerate training by randomly selecting a subset of layers. 
In addition, \citet{yao2024layer} utilizes a reinforcement learning strategy to dynamically select the most important subsets with layer-wise importance scoring, achieving better performance and reducing memory as well. However, they typically treat all data points uniformly and apply the same subsets to every input regardless of its complexity or relevance, which limits their ability to exploit the diverse learning signals present across the training corpus.

\subsection{Data-wise Sparse Tuning}
Recently, increasing attention has been directed toward developing methods for selecting data before training foundation models. \citet{xia2024less} introduces an optimizer-aware algorithm that estimates data influence under the Adam optimizer and performs low-rank gradient similarity search to select instruction data. Other methods investigate online batch selection to enhance the training of LLMs based on gradient norm or maximum sample loss of ``hard samples'' \cite{katharopoulos2018not,jiang2019accelerating}, or explore using additional reference models to more accurately estimate the importance of samples \cite{deng2023towards}. Furthermore, \citet{wang2024greats} applies a greedy algorithm to optimize the data batch quality approximated by Taylor expansion, efficiently capturing true data informativeness. However, they still treat layer selection uniformly across inputs, which may cause gradient conflicts or insufficient information capturing for specific layers.

\section{Method}
In this section, we first present our theoretical motivation and show that the alignment of training gradients with a support gradient can guide more effective parameter updates. Next, we introduce our method, Gradient-aligned Sparse Tuning (GAST). GAST dynamically selects adapter parameters that are most positively aligned with the support gradient, improving performance and generalization.

\subsection{Theoretical Motivation}
We consider a PEFT-equipped LLM as $\mathcal{M}_{\Theta+\Delta}$, where $\Theta$ are the frozen backbone weights and $\Delta$ are the trainable adapter parameters. Given a small support set $\mathcal{D}_\text{sup}$, the gradient induced by the support set at step $t$ for layer $i$ is $g_{t,\text{sup}}^{(i)}=\mathbb{E}_{\mathcal{D}_\text{sup}}\nabla \ell(\Theta,\Delta_t;\mathcal{D}_\text{sup})$. The gradient induced by a training example $x_j$ is $g_{t,j}^{(i)}=\nabla \ell(\Theta,\Delta_t;x_j)$. Consider the gradient alignment between training gradients and support gradients:
\begin{equation}
    \mathrm{sim}(g_{t,j}^{(i)},g_{t,\text{sup}}^{(i)})=\frac{\langle g_{t,j}^{(i)},g_{t,\text{sup}}^{(i)} \rangle}{||g_{t,j}^{(i)}|| \, ||g_{t,\text{sup}}^{(i)}||}.
    \nonumber
\end{equation}
A positive alignment implies that updating parameters using sample $j$ will effectively reduce support-set loss, while a negative alignment indicates gradient conflict as shown in Fig.~\ref{fig:overall}. For a layer-wise spare tuning strategy, it uses all training data on the selected layers. We define its gradient aggregation for layer $i$ as: $g_{t,{\text{layer}}}^{(i)}={\frac{1}{|\mathcal{D}_\text{train}|}}\sum_{j\in {\mathcal{D}_\text{train}}}g_{t,j}^{(i)}$. Similarly, for the data-wise selection strategy, we have the following formula: $g_{t,{\text{data}}}^{(i)}={\frac{1}{|\mathcal{D}_\text{sub}|}}\sum_{j\in {\mathcal{D}_\text{sub}}}g_{t,j}^{(i)}$, where $\mathcal{D}_\text{sub}$ is the selected subset of the training data for each layer. To leverage the superiority of our hybrid data-layer selective sparse tuning, we define the dynamically selected subset at layer $i$ as: $\mathcal{D}_+^{(i)}=\{ x_j:\langle g_{t,j}^{(i)},g_{t,\text{sup}}^{(i)} \rangle > 0\}$. Thus the aggregated gradient at layer $i$ is: $g_{t,\text{hybrid}}^{(i)}={\frac{1}{|\mathcal{D}_+^{(i)}|}}\sum_{j\in \mathcal{D}_+^{(i)}}g_{t,j}^{(i)}$.

Consider the magnitude of gradient projection onto the support gradient $\langle g_{t,\cdot}^{(i)},g_{t,\text{sup}}^{(i)} \rangle$, there exists one example $x_{j'}\in \mathcal{D}_\text{sub}, \quad \mathrm{s.t.}\quad \langle g_{t,j'}^{(i)},g_{t,\text{sup}}^{(i)} \rangle < 0$ for the data-selective strategy, which selected fixed subset for the training data. For the layer-selective strategy using all training data, there always exists a gradient conflict having a negative gradient alignment. However, the hybrid selection achieves strictly greater effective gradient magnitude toward support-set minimization with each term $\langle g_{t,j}^{(i)},g_{t,\text{sup}}^{(i)} \rangle > 0$ for $x_j \in \mathcal{D}_+^{(i)}$.
With the assumptions that 1) a non-empty set of positively aligned per-sample gradients;
2) negative-alignment gradients do not dominate in magnitude, and
3) the baseline data subset $\mathcal{D}_\text{sub}$ is fixed and not constructed using gradient alignment. Thus, we could have:
\begin{equation}
\small
    \langle g_{t,\text{hybrid}}^{(i)},g_{t,\text{sup}}^{(i)} \rangle \geq \text{max}\{ \langle g_{t,{\text{layer}}}^{(i)},g_{t,\text{sup}}^{(i)} \rangle, \langle g_{t,{\text{data}}}^{(i)},g_{t,\text{sup}}^{(i)} \rangle \}.
    \label{gali}
\end{equation}

\newtheorem{lemma}{Lemma}
Consider a smoothness assumption on the loss function (detailed proof in the supplementary):
\begin{lemma}[L-Smoothness]\label{lemma1}
Let $\ell(\Delta)$ be an $L$-smooth objective with respect to $\Delta$. At iteration $t$, let $\Delta_t$ denote the current parameters and let $g_t$ be a stochastic gradient estimator satisfying
\begin{equation}
\mathbb{E}[g_t \mid \Delta_t] = \nabla \ell(\Delta_t).
\end{equation}
With step size $\eta_t > 0$, the conditional expectation of the loss satisfies
\begin{equation}
\small
\begin{aligned}
\mathbb{E}[\ell(\Delta_{t+1}) \mid \Delta_t]
\;\le\;&
\ell(\Delta_t)
- \eta_t\, \mathbb{E}\!\left[
  \big\langle \nabla \ell(\Delta_t),\, g_t \big\rangle
  \,\middle|\, \Delta_t
\right] \\
&+ \frac{L \eta_t^{2}}{2}\,
\mathbb{E}\!\left[ \|g_t\|^{2} \,\middle|\, \Delta_t \right].
\end{aligned}
\end{equation}
Here $C>0$ is the Lipschitz constant. In particular, for fixed $\eta_t$ and bounded 
$\mathbb{E}[\|g_t\|^{2} \mid \Delta_t]$, any strategy that yields a larger 
$\mathbb{E}\!\left[
 \langle \nabla \ell(\Delta_t),\, g_t \rangle
 \,\middle|\, \Delta_t
\right]$
(i.e., better alignment with the true gradient) leads to a larger expected one-step decrease in the loss.
\end{lemma}

Since our method maximizes the projection $\langle \nabla\ell(\Delta_{t}),g \rangle$, we obtain the fastest decrease in loss per step given the same support set and learning rate. For each layer $i$, the expected reduction in loss is largest for our hybrid strategy based on Eq.~\ref{gali}, due to greater gradient alignment and magnitude. Formally, the per-step expected loss reduction satisfies:
\begin{equation}
\small
    \ell(\Delta_{{t+1},{\text{hybrid}}}^{(i)}) \leq \min \{ \ell(\Delta_{{t+1},\text{layer}}^{(i)}),\ell(\Delta_{{t+1},\text{data}}^{(i)}) \}.
    \label{aliloss}
\end{equation}

\subsection{Gradient-aligned Sparse Tuning}

Derived from Eq.~\ref{aliloss}, under such \emph{gradient heterogeneity}, both traditional \emph{layer-selective} and \emph{data-selective} strategies can only achieve sub–optimal solutions when the batch size is larger than one. To overcome this limitation, as shown in Fig.~\ref{fig:overall}, we introduce GAST, which dynamically assigns different adapter layers to different training examples within the same mini-batch, based on their instantaneous gradient information.
Concretely, for every mini-batch, we compare the gradients of the training samples with the gradients obtained on a held-out support set, and select the layers whose gradients are the most similar—the whole procedure is performed on-the-fly.

Starting from an initial adapter $\Delta_0$, standard full-adapter training proceeds as
\begin{equation}
    \Delta_{t+1} = \Delta_t-\eta_t\sum_{x\in \mathcal{B}_t}\nabla \ell(\Theta,\Delta_t;x),
\end{equation}
where $\mathcal{B}_t$ is the mini-batch selected at step $t$ and $\eta_t$ is the learning rate. Instead of updating all adapter parameters for all samples, GAST seeks, for every example $x_j\in \mathcal{B}_t$, a sample-specific subset of adapter parameters $\widehat\Delta_{t,j}\in\Delta_t$, such that only the weights in $\widehat\Delta_{t,j}$ are updated with gradient from $x_j$.


Let $\mathcal{X}_{\text{train}}$ and $\mathcal{X}_{\text{sup}}$ denote the training and support sets, respectively. 
\newtheorem{theorem}{Theorem}
Considering the total differential theory related to the first-order Taylor expansion \cite{xia2024less} (detailed in the Supplementary):
\begin{theorem}[Total Differential.]\label{theorem1}
The sum of the products of each partial derivative and the corresponding small change in the weight variable can estimate the increment in the loss function at a given point.
\end{theorem}
Let $w^{(i)}$ be the adapter weights of layer $i \in \{1,\dots,L\}$, and let $\delta^{(i)}$ be the small change applied to $w^{(i)}$.
Then the increment of the support-set loss obeys
\begin{equation}
    \mathcal{L}_{t+1} = \mathcal{L}_t + \sum_{i\in \{1,...,L\}} \left \langle\frac{\partial\mathcal{L}_t}{\partial w^{(i)}},\delta^{(i)}\right \rangle ,
    \label{eq:1}
\end{equation}
where $\mathcal{L}_t=\sum_{x_k\in\mathcal{X}_{\text{sup}}}\ell(\Theta,\Delta_t;x_k)$ denote the support-set loss, $\left\langle\cdot,\cdot\right\rangle$ denotes inner product. 

\begin{algorithm}[tb]
\caption{Gradient-aligned Sparse Tuning (GAST)}
\label{alg:rl} 
\small
\begin{algorithmic}[1] 
\REQUIRE LLM $\mathcal{M}$ with $L$ layers equipped with PEFT parameterized by $\Delta$, training dataset $\mathcal{D_{\text{train}}}$, support dataset $\mathcal{D_{\text{sup}}}$
\STATE Initialize PEFT parameters $\Delta_0$ for all layers.
\FORALL{$t = 0,\dots,T-1$}
\STATE Sampled a random mini-batch $\mathcal{B}_t\sim\mathcal{D_{\text{train}}}$
\FORALL{$i = 0,\dots,L-1$}
\FORALL{$j = 0,\dots,|\mathcal{B}_t|-1$}
\STATE Calculate gradient alignment score $s_{t,j}^{(i)}=\left\langle g_{t,\text{\text{sup}} }^{(i)},g_{t,j}^{(i)}\right\rangle$
\ENDFOR
\STATE Calculate the sampling probability $p_{t, j}^{(i)}$ (see Eq.~\ref{eq:4})
\STATE Sampling the updating indices $j^{*}(i)$ (see Eq.~\ref{eq:5})
\STATE Update PEFT parameters $\Delta_{t+1}^{(i)}=\Delta_{t}^{(i)}-\eta_t\nabla g^{(i)}_{t,(j^*(i))}$
\ENDFOR
\ENDFOR
\end{algorithmic}
\end{algorithm}

For a mini-batch $\mathcal{B}_t = \{x_{t,1},\dots,x_{t,|\mathcal{B}_t|}\}$, for every layer $i$ and every sample $x_{t,j}\in\mathcal{B}_t$, we have the layer-wise, sample-specific gradient: 
\begin{equation}
g_{t,j}^{(i)}=\nabla_{\Delta^{(i)}}\ell(\Theta,\Delta_t;x_{t,j}).
    \label{eq:2}
\end{equation}
Hence, setting $\delta^{(i)} = -\eta_t g_{t,j}^{(i)}$ and 
$g_{\text{sup}}^{(i)}=\nabla_{w^{(i)}}\mathcal{L}_t={\partial\mathcal{L}_t}/{\partial w_i}$ we obtain
\begin{equation}
    \mathcal{L}_{t+1} -\mathcal{L}_t =  - \eta_t \sum_{i=1}^L s_{t,j}^{(i)}, \quad s_{t,j}^{(i)}=\left\langle g_{t,\text{\text{sup}} }^{(i)},g_{t,j}^{(i)}\right\rangle,
\end{equation}
where 
$s_{t,j}^{(i)}$ denote the gradient alignment score. Note that computing $g_{t,\text{sup}}$ over the support set at every step should be prohibitively expensive. Therefore, we sample a small subset of support data at each iteration, with a batch size much smaller than that of the training batch.

Eq.~\ref{eq:1} suggests that, for every layer $i$, we should pick the training example whose gradient aligns most positively with the current support-set gradient in order to maximize the expected decrease in support-set loss. 

To mitigate the risk of overfitting to the support set, we employ a stochastic selection scheme. Specifically, we first compute the sampling probability for each sample in the batch based on the normalized alignment score:
\begin{equation}
    p_{t, j}^{(i)} = \frac{\exp(\widehat{s}_{t, j}^{(i)})}{\sum_k \exp(\widehat{s}_{t, k}^{(i)})},\quad \widehat{s}_{t,\cdot}^{(i)} = \mathrm{Norm}({s}_{t,\cdot}^{(i)})
\label{eq:4}
\end{equation}
where $\mathrm{Norm}(\cdot)$ denotes mean-std normalization over mini-batch.

Next, for each layer $i$, we sample $K$ indices according to $p_{t, j}^{(i)}$, and denote the index used to update layer $i$ as
\begin{equation}
j^{*}(i)\sim \text{Categorical}\bigl(p_{t,1}^{(i)},\dots,p_{t,|\mathcal{B}_t|}^{(i)}\bigr).
\label{eq:5}
\end{equation}

To achieve data-layer selective updating, only the gradient of the selected data points $x_{t, j^*(i)}$ are used to update layer $i$, i.e.,
\begin{equation}
    \Delta_{t+1}^{(i)}=\Delta_{t}^{(i)}-\eta_t g^{(i)}_{t,j^*(i)}.
\end{equation}
Intuitively, GAST enables each adapter layer to learn only from the most relevant training examples in a mini-batch, thereby making parameter updates more targeted and effective. This data-layer-selective sparse tuning both leverages gradient heterogeneity and promotes better generalization. We detail the complete GAST training procedure in Algorithm~1 for better clarification.



\section{Experiments}
\subsection{Experimental Setup}
\paragraph{Models.} We selected LLaMA-7B, LLaMA-13B~\cite{llama}, GPT-J-6B~\cite{gpt-j}, and LLaMA3-8B as the foundational models for downstream task finetuning, considering their widespread use in the research community. For reference, we also report results from ChatGPT (gpt-3.5-turbo) using zero-shot Chain-of-Thought prompting~\cite{wei2022chain}.

\paragraph{Datasets.} We evaluated our method on commonsense and arithmetic reasoning tasks.
For commonsense reasoning, we utilized eight sub-tasks, each with standard training and testing splits: BoolQ~\cite{boolq}, PIQA~\cite{piqa}, SIQA~\cite{siqa}, HellaSwag~\cite{hellaswag}, WinoGrande~\cite{winogrande}, ARC~\cite{arc}, and OBQA~\cite{openbookqa}. Following \citet{llmadapter}, we aggregated the training sets of all these tasks for joint training and evaluated the fine-tuned models on the official test split for each sub-task.
For arithmetic reasoning, we fine-tuned models on the Math10K dataset, which contains math reasoning samples curated by \citet{llmadapter}. Evaluation was performed on the official test sets of datasets, including GSM8K~\cite{gsm8k}, AQuA~\cite{aqua}, MAWPS~\cite{mawps}, and SVAMP~\cite{svamp}.

\paragraph{Implementation Details.}
Our experiments are conducted using the LLM-Adapter framework~\cite{llmadapter}. Consistent with prior work, we set the batch size to 16 and the number of training epochs to 3. In GAST, we treat the whole training set as the support set; at every iteration we randomly pick four samples from it to calculate the averaged support-set gradient, and set $K=8$. For PEFT, we adopt Series Adapter~\cite{series}, Parallel Adapter~\cite{parallel}, and Low-rank Adapter (LoRA)~\cite{lora}. For LoRA, we used a rank of 32 with an alpha of 64, a dropout rate of 0.05, and applied it to the {Q, K, V, Up, Down} modules. For the Series and Parallel adapters, we employed a bottleneck size of 256, equipping them with {Up, Down} and {Up, Gate} modules, respectively. A uniform learning rate of 1e-4 with 100 warmup steps was set for all methods, and a maximum context length of 256 was selected.
For the 7B models, training was conducted for 11.5 hours. All experiments were conducted on a workstation equipped with an NVIDIA A100 80GB GPU.

\subsection{Baselines} 
We compare GAST with the following advanced representative adaptive methods:\\
\textbf{LISA}~\cite{pan2024lisa}: A PEFT approach that sparsely tunes only a single transformer layer, significantly reducing trainable parameters.\\
\textbf{AdaLoRA}~\cite{zhang2023adaptive}: A rank-adaptive PEFT method that can be broadly applied to various reparameterization-based strategies, which dynamically adjusts the rank size of each LoRA.\\ 
\textbf{RST}~\cite{yao2024layer}: This method randomly selects and fine-tunes a subset of transformer layers, thereby reducing trainable parameters.\\ 
\textbf{IST}~\cite{yao2024layer}: The latest layer-selective sparse tuning method, which employs reinforcement learning to rank layer importance.\\ 
\textbf{GREATS}~\cite{wang2024greats}: A recent online data selection approach that utilizes the calculation of Data Shapley~\cite{wang2024data} values in an online manner. At each step, GREATS selects the optimal half of the training data greedily. 

\begin{table}[t]
\small
\centering
\begin{tabular}{ccc}
\toprule
Method & Adaptiveness & Average \\\midrule
LoRA & - & 74.7 (+0.0)\\
LISA & Layer & 75.3 (+0.6)\\
AdaLoRA & Rank & 76.2 (+1.5)\\
LoRA + RST & Layer & 75.8 (+1.1)\\
LoRA + IST & Layer & 76.5 (+1.8)\\
LoRA + GREATS & Data & 76.3 (+1.6)\\
\rowcolor{mygray} LoRA + GAST & Data \& Layer & \textbf{77.5 (+2.8)}\\\bottomrule
\end{tabular}
\caption{Comparison with different adaptive methods.}
\label{tab:adapt}
\end{table}

\begin{figure}[!t]
\centering
\includegraphics[width=.85\columnwidth]{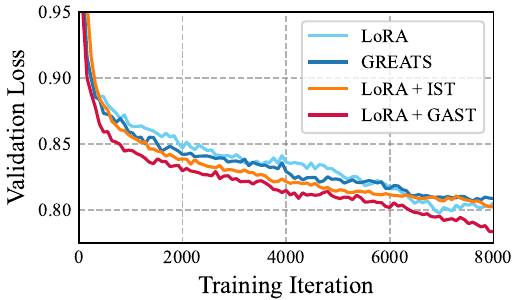}
\caption{Comparison of model convergence with loss curve.}
\label{fig:comp} 
\end{figure}

\begin{table*}[t]
\setlength{\tabcolsep}{1.2mm}
\centering
\resizebox{0.99\textwidth}{!}{
\begin{tabular}{clcccccccc|cc}
\toprule
\textbf{Model} & \multicolumn{1}{c}{\textbf{PEFT}}  & \textbf{BoolQ} & \textbf{PIQA}&\textbf{SIQA}& \textbf{HellaSwag} & \textbf{WinoGrande}& \textbf{ARC-e} & \textbf{ARC-c} & \textbf{OBQA} & \textbf{Avg.} \\ \hline
ChatGPT&\multicolumn{1}{c}{-}&73.1&85.4&68.5&78.5&66.1&89.8&79.9&74.8&77.0\\\hline
\multirow{9}{*}{LLaMA$_{\text{7B}}$}&Series&63.0&79.2&76.3&67.9&75.7&74.5&57.1&72.4&70.8\\
&Series + IST&66.2&78.3&74.9&72.2&75.9&75.8&59.0&72.2&71.8\\
 &\cellcolor{mygray}Series + GAST&\cellcolor{mygray}68.0 &\cellcolor{mygray}79.1&\cellcolor{mygray}77.8&\cellcolor{mygray}79.8&\cellcolor{mygray}78.5 &\cellcolor{mygray}77.4 &\cellcolor{mygray}62.0&\cellcolor{mygray}74.8&\cellcolor{mygray}\textbf{74.7}\\
&Parallel&67.9&76.4&78.8&69.8&78.9&73.7&57.3&75.2&72.2\\
&Parallel + IST&68.4&79.1&77.9&70.0&78.9&81.2&62.3&77.6&74.4\\
 &\cellcolor{mygray}Parallel + GAST&\cellcolor{mygray}67.9 &\cellcolor{mygray}81.3 &\cellcolor{mygray}78.4 &\cellcolor{mygray}80.5 &\cellcolor{mygray}79.8 &\cellcolor{mygray}78.2 &\cellcolor{mygray}63.0 &\cellcolor{mygray}77.4 &\cellcolor{mygray}\textbf{75.8} \\
&LoRA&68.9&80.7&77.4&78.1&78.8&77.8&61.3&74.8&74.7\\
&LoRA + IST&68.7&81.7&77.3&82.7&78.7&80.6&62.4&80.0&76.5\\
 &\cellcolor{mygray}LoRA + GAST&\cellcolor{mygray}68.2 &\cellcolor{mygray}81.6&\cellcolor{mygray}79.4 &\cellcolor{mygray}83.6 &\cellcolor{mygray}82.2 &\cellcolor{mygray}80.4 &\cellcolor{mygray}64.7 &\cellcolor{mygray}79.8 &\cellcolor{mygray}\textbf{77.5} \\\hline
\multirow{9}{*}{LLaMA$_{\text{13B}}$}&Series&71.8&83.0&79.2&88.1&82.4&82.5&67.3&81.8&79.5\\
&Series + IST&  72.9&82.2&81.4&87.9&84.0&82.7&69.1&81.1&80.2\\
 &\cellcolor{mygray}Series + GAST&\cellcolor{mygray}72.9 &\cellcolor{mygray}82.0  &\cellcolor{mygray}82.8 &\cellcolor{mygray}89.5  &\cellcolor{mygray}85.8&\cellcolor{mygray}84.2 &\cellcolor{mygray}70.6  &\cellcolor{mygray}82.0 &\cellcolor{mygray}\textbf{81.2} \\
&Parallel&72.5&84.9&79.8&92.1&84.7&84.2&71.2&82.4&81.4\\
&Parallel + IST&72.6&86.0&79.2&89.1&83.5&84.8&70.6&82.8&81.1\\
 &\cellcolor{mygray}Parallel + GAST&\cellcolor{mygray}72.8 &\cellcolor{mygray}86.1 &\cellcolor{mygray}80.6 &\cellcolor{mygray}91.0 &\cellcolor{mygray}85.8 &\cellcolor{mygray}86.0 &\cellcolor{mygray}72.1 &\cellcolor{mygray}84.0 &\cellcolor{mygray}\textbf{82.3} \\
&LoRA&72.1&83.5&80.5&90.5&83.7&82.8&68.3&82.4&80.5\\
&LoRA + IST&71.5&85.0&81.2&89.1&84.2&84.0&70.1&81.8&80.9\\
 &\cellcolor{mygray}LoRA + GAST&\cellcolor{mygray} 73.2&\cellcolor{mygray}84.5 &\cellcolor{mygray}81.9 &\cellcolor{mygray}90.8 &\cellcolor{mygray}85.5 &\cellcolor{mygray}84.6 &\cellcolor{mygray}71.5 &\cellcolor{mygray}82.9 &\cellcolor{mygray}\textbf{81.8} \\\hline\hline
\multirow{3}{*}{GPT-J$_{\text{6B}}$}&LoRA&62.4&68.6&49.5&43.1&57.3&43.4&31.0&46.6&50.2\\
&LoRA + IST&63.0&63.2&62.9&35.8&39.1&56.8&39.1&51.2&51.4\\
 &\cellcolor{mygray}LoRA + GAST&\cellcolor{mygray}63.1 &\cellcolor{mygray}74.4 &\cellcolor{mygray}65.0 &\cellcolor{mygray}49.7 &\cellcolor{mygray}59.9 &\cellcolor{mygray}59.7 &\cellcolor{mygray}45.3 &\cellcolor{mygray}60.2 &\cellcolor{mygray}\textbf{59.7} \\\hline\multirow{3}{*}{LLaMA3$_{\text{8B}}$}&LoRA&70.8&85.2&79.9&91.7&84.3&84.2&71.2&79.0&80.8\\
&LoRA + IST&72.7&88.3&80.5&94.7&84.4&89.8&79.9&86.6&84.6\\
 &\cellcolor{mygray}LoRA + GAST&\cellcolor{mygray}73.6 &\cellcolor{mygray}87.4 &\cellcolor{mygray}81.0 &\cellcolor{mygray}95.2 &\cellcolor{mygray}86.5 &\cellcolor{mygray}90.0 &\cellcolor{mygray}79.8 &\cellcolor{mygray}85.0 &\cellcolor{mygray}\textbf{84.8} \\
\bottomrule
\end{tabular}}
\vspace{-2pt}         
\caption{Comparison across multiple LLMs using different PEFT approaches on eight commonsense reasoning benchmarks. Baseline results for GPT-J and LLaMA are taken from \citet{llmadapter}.
}
\label{tab:llama_commonsense}
\vspace{-8pt}
\end{table*}

\begin{table}[t]
\small
\setlength{\tabcolsep}{0.8mm}
\centering
\begin{tabular}{lcccccc}
\toprule
 \multicolumn{1}{c}{\textbf{Method}}  & \textbf{GSM8K} & \textbf{AQuA}&\textbf{MAWPS}& \textbf{SVAMP}& \textbf{Avg.} \\ \hline
\multicolumn{1}{c}{ChatGPT}&56.4&38.9&87.4&69.9&63.2\\\hline
LoRA&61.0&26.4&91.6&74.4&63.4\\
LoRA + IST&62.8&31.5&89.9&76.3&64.7 \\
\rowcolor{mygray} LoRA + GAST& 66.4 & 32.7 & 91.6 & 79.4& \textbf{67.5}\\
\bottomrule
\end{tabular}
\vspace{-2pt}
\caption{Comparison with LLaMA3-8B on four math reasoning datasets.}
\label{tab:llama_math}
\vspace{-8pt}
\end{table}
\begin{table}[t]
\small
\centering
\begin{tabular}{lcc}\toprule
Setting &Data-layer Selection&Average \\\midrule
LoRA &  - & 74.7\\
LoRA + GAST & Random Selection & 76.4 \\
LoRA + GAST & Top-k Selection & 66.4 \\
LoRA + GAST & Sampling-based Selection &\textbf{77.5} \\\bottomrule
\end{tabular}
\caption{The effect of data-layer selection strategy.}
\label{tab:ablate}
\end{table}

\subsection{Comparison with Adaptive Methods}
The comparison results shown in Tab.~\ref{tab:adapt} demonstrate the consistent improvement and effectiveness of our GAST method with other adaptive methods on the commonsense task evaluated on the LLaMA 7B model. The standard LoRA baseline achieves an average score of 74.7, while methods utilizing adaptiveness at either the layer or rank level, such as LISA, RST, and AdaLoRA, achieve promising improvements, with AdaLoRA reaching 76.2 by dynamically adjusting module ranks. Data-selective methods like GREATS also provide notable gains (76.3), and layer-selective methods such as IST further enhance prediction performance (76.5). Among all compared approaches, GAST achieves the highest average score (77.5), indicating that jointly leveraging gradient heterogeneity across both data and layers results in more effective parameter tuning.

To further compare the methods, we visualized the validation loss of different approaches to compare the convergence of the models. As shown in Fig.~\ref{fig:comp}, the data-selective GREATS and layer-selective IST methods significantly outperform the baseline LoRA in the early stages of training but experience fluctuations in the middle stages. This may be due to gradient conflict on data for each layer, which manifests in the middle stages of the model. In contrast, our method consistently surpasses all other methods, demonstrating its effectiveness in overcoming gradient conflict.
 
\subsection{Evaluation of Versatility} 
To further demonstrate the broad improvements provided by our method for various PEFT approaches, we conducted experiments on different models, datasets, and PEFT methods. The quantitative results in Tab.~\ref{tab:llama_commonsense} present an extensive evaluation of the effectiveness of GAST across a wide range of PEFT settings and LLM backbones. We observe that integrating GAST consistently enhances the performance of various models on commonsense reasoning benchmarks. Taking LLaMA-7B as an example, GAST leads to clear accuracy improvements under all three PEFT strategies (Series, Parallel, and LoRA). For instance, on the challenging HellaSwag dataset, LoRA+GAST improves accuracy from 78.1\% to 83.6\% compared to standard LoRA, representing a notable advance. This trend holds on other competitive datasets such as WinoGrande and ARC-c as well. Additionally, GAST brings significant gains to GPT-J-6B, where the average accuracy increases by over 9 points over the LoRA baseline. These improvements are consistent across almost all datasets and model configurations, confirming the robustness and broad applicability. 

We further assess GAST’s effectiveness on a series of math reasoning benchmarks, as summarized in Tab.~\ref{tab:llama_math}. GAST demonstrates consistent and notable improvements when compared to IST. For example, when applying GAST on LLaMA3-8B with LoRA, we see increases not only in the average score (from 63.4\% with LoRA to 67.5\% with LoRA+GAST), but also robust improvements on individual datasets such as GSM8K and SVAMP. These results indicate that GAST  is not only beneficial for commonsense reasoning, but also generalizes well to tasks requiring complex mathematical reasoning. The observed gains across such diverse benchmarks further highlight the versatility and practical value of GAST in enriching parameter-efficient fine-tuning of large language models.
  
\subsection{Analytical Studies}
\paragraph{Ablation Study.}
We conducted experiments to evaluate the effects of gradient-aligned sparse tuning by training the LLaMA 7B model with LoRA on a commonsense task and reporting the average accuracy. As shown in Tab.~\ref{tab:ablate}, using random data-layer selection led to significant performance improvements, consistent with the findings of IST~\cite{yao2024layer} that training with fewer parameters can enhance generalization. However, adopting a top-k selection strategy did not improve performance. 
This may be due to the small mini-batch drawn from the support set, which fails to capture the overall gradient distribution of the test set.
Finally, we employed a sampling-based selection approach, which yielded the best results and further demonstrated the effectiveness of GAST.

\begin{figure}[t]
\centering
\begin{subfigure}[b]{0.49\linewidth}
    \centering
    \includegraphics[width=\linewidth]{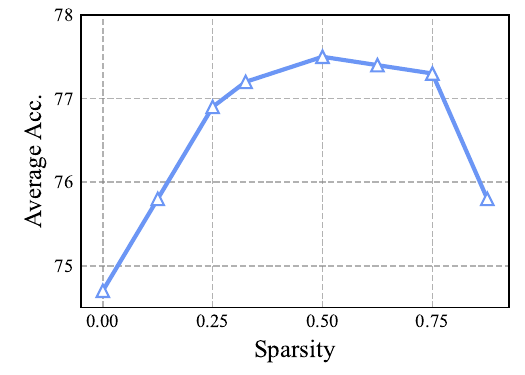}
    \captionsetup{aboveskip=0pt, belowskip=0pt}
    \caption{Data-layer Sparsity}
\end{subfigure}%
\begin{subfigure}[b]{0.49\linewidth}
    \centering
    \includegraphics[width=\linewidth]{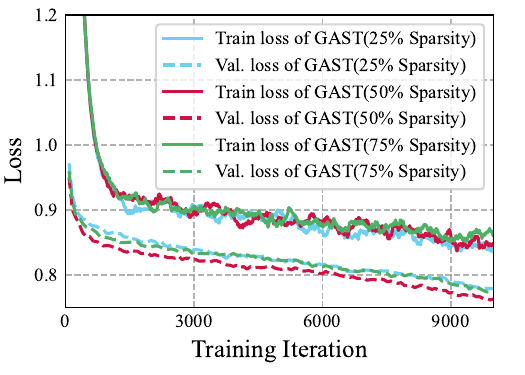}
    \captionsetup{aboveskip=0pt, belowskip=0pt}
    \caption{Loss curve}
\end{subfigure}
\caption{Impact of data-layer sparsity in GAST.}
\label{fig:sparsity}
\end{figure}
\begin{figure}[t]
\centering
\begin{subfigure}[b]{0.49\linewidth}
    \centering
    \includegraphics[width=\linewidth]{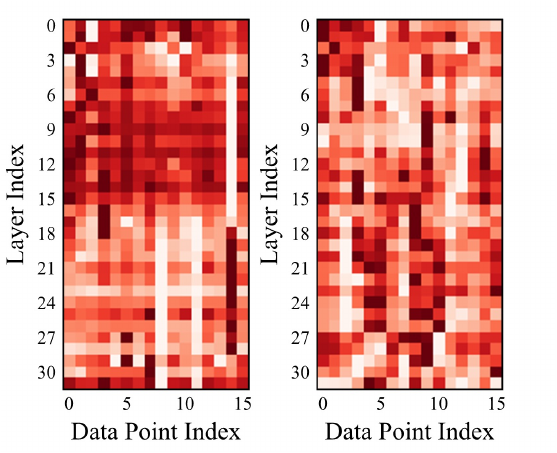}
    \captionsetup{aboveskip=0pt, belowskip=0pt}
    \caption{Sampling probability}
\end{subfigure}%
\begin{subfigure}[b]{0.49\linewidth}
    \centering
    \includegraphics[width=\linewidth]{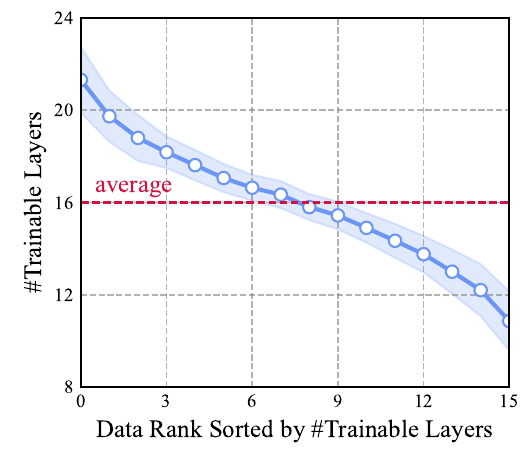}
    \captionsetup{aboveskip=0pt, belowskip=0pt}
    \caption{Num. of sampled layers}
\end{subfigure}
\caption{(a) Visualization of the sampling probability of mini-batch data points across layers on two different iterations. A deeper red color indicates a higher probability.
(b) Distribution of the number of layers each data point is trained on within a mini-batch.}
\label{fig:sampling} 
\end{figure}

\paragraph{Impact of Sparsity in GAST.}
We conducted experiments to evaluate the effect of data-layer sparsity in GAST by training the LLaMA 7B model with LoRA on a commonsense reasoning task. As shown in Fig.~\ref{fig:sparsity}(a), we gradually increased the data-layer sparsity from 0.0 to 0.875, reporting the averaged test accuracies. Here, a sparsity of 0.0 corresponds to the standard LoRA setting, while 0.5 is our default configuration. The experimental results indicate that when the sparsity is set to an extremely high value (0.875), the network becomes overly sparse and performance drops. Nevertheless, even in this highly sparse regime, the model still outperforms the baseline standard LoRA, which suggests the effectiveness of data-layer selective utilization. Between sparsity values of 0.25 and 0.75, the performance remains relatively stable, reaching its peak at 0.5, after which it gradually declines. Fig.~\ref{fig:sparsity}(b) shows the training and validation loss curves corresponding to different sparsity levels. It can be observed that the denser setting (i.e., with 75\% sparsity) exhibits a faster decrease in training loss during the initial stage. However, both excessively high and low sparsity levels fail to yield a lower validation loss. This may be because low sparsity can cause gradient conflicts, while high sparsity results in insufficient information being retained.
These results suggest that a sparsity level of 0.5 strikes a balance between preserving important information and avoiding gradient conflicts.

\paragraph{Distribution of Sampled Layers.}
To gain a deeper understanding of our method, we visualize the sampling probabilities of 16 data points within a mini-batch, across the 32 layers of LLaMA-7B. A higher sampling probability indicates a higher gradient alignment score. As shown in Fig~\ref{fig:sampling}(a), different data points exhibit distinct probability distributions across the layers. For instance, in the left figure, the 8th and 11th data points have higher probabilities in the shallow layers, whereas the 14th data point shows higher probabilities in the deeper layers. This demonstrates that our approach can effectively estimate the contribution of each data point to different layers.
Next, we visualize the number of sampled layers for each data point per iteration. As shown in Fig~\ref{fig:sampling}(b), although the sparsity is set to 50\%, the most important data points are trained in up to 70\% of the layers, while the least important ones are only trained in 30\% of the layers. This indicates that our GAST dynamically allocates the number of layers fine-tuned for each data point according to its gradient alignment. 

\section{Conclusion}
In this work, we introduced GAST, a novel PEFT framework that jointly considers both data point and layer selection to finetune LLMs. The proposed GAST dynamically assigns informative data to specific layers based on gradient alignment with a holdout support set.
This fine-grained, data-layer selection mechanism effectively mitigates gradient conflicts, improving both convergence and performance. We theoretically demonstrated that our hybrid selection strategy yields strictly better gradient alignment and faster convergence compared to existing layer-wise or data-wise sparse tuning methods. 
We believe this work opens promising avenues for more adaptive and generalizable tuning paradigms for large models.

\section*{Limitations}
There are two limitations in this work. First, similar to GREATS, our method cannot simultaneously reduce both memory usage and computational cost due to engineering optimization constraints, even though it could potentially achieve higher performance. Second, because of limited resources, we were unable to validate larger language models such as LLaMA 3 70B. Whether larger models require sparser fine-tuning remains unknown, and we leave this as future work.

\bibliography{custom}
\clearpage
\appendix

\section{More Details of Our Method}
\subsection{Proof of L-Smoothness}
\newtheorem{assumption}{Assumption}
Consider the loss function $\ell(\cdot)$, we have the following assumption:
\begin{assumption}\label{assump1}
\textbf{(Lipschitz-continuous objective gradients)}\\
Let $\ell$ be a continuously differentiable loss function and the gradient function of $\ell$ as $\nabla \ell$ is Lipschitz continuous with Lipschitz constant $L>0$, i.e.,
\begin{equation}
\begin{aligned}
    ||\nabla \ell(\Delta_{t+1})-\nabla \ell(\Delta_{t})||_2 & \leq L||\Delta_{t+1}-\Delta_{t}||_2 \\
    &for \  all \, \{ \Delta_{t+1},\Delta_{t}\} \subset \mathbb{R}^d.
\end{aligned}
\end{equation}
\end{assumption}
This assumption ensures that the gradient of $\ell$ does not change arbitrarily quickly with respect to the parameter vector. Then we could have the following formular:
\begin{equation}
    \begin{aligned}
        \ell(\Delta_{t+1}) \leq & \ell(\Delta_{t}) + \nabla\ell(\Delta_{t})^T(\Delta_{t+1}-\Delta_{t}) + \\
        &\frac{1}{2}L||\Delta_{t+1}-\Delta_{t}||^2_2\\
        & for \ all \quad \{\Delta_{t+1},\Delta_{t}\} \subset \mathbb{R}^d. 
    \end{aligned}
    \label{inequ2}
\end{equation}
Under Assumption 1, we can obtain:
\begin{equation}
\small
\begin{aligned}
    &\ell(\Delta_{t+1})=\ell(\Delta_{t}) + \int_0^1 \frac{\partial\ell(\Delta_{t} + s(\Delta_{t+1}-\Delta_{t}))}{\partial s} \, ds \\
    &=\ell(\Delta_{t}) + \int_0^1 \nabla\ell(\Delta_{t} + s(\Delta_{t+1}-\Delta_{t}))^T(\Delta_{t+1}-\Delta_{t}) \, ds \\
    &=\ell(\Delta_{t}) + \nabla\ell(\Delta_{t})^T(\Delta_{t+1}-\Delta_{t})) + \\
    &\int_0^1[\nabla\ell(\Delta_{t} + s(\Delta_{t+1}-\Delta_{t})) - \nabla\ell(\Delta_{t})]^T(\Delta_{t+1}-\Delta_{t}) \, ds \\
    &\leq \ell(\Delta_{t}) + \nabla \ell(\Delta_{t})^T(\Delta_{t+1}-\Delta_{t}) + \\
    & \int_0^1 L||s(\Delta_{t+1}-\Delta_{t})||_2||\Delta_{t+1}-\Delta_{t}||_2 \, ds,
\end{aligned}
\end{equation}
which results in the inequality \ref{inequ2}. Then, according to the iterates of Stochastic Gradient, we can have:
\begin{equation}
\small
    \begin{aligned}
        \ell(\Delta_{t+1}) - \ell(\Delta_{t}) &\leq \nabla\ell(\Delta_{t})^T(\Delta_{t+1}-\Delta_{t}) \\ 
        & \quad + \frac{1}{2}L||\Delta_{t+1}-\Delta_{t}||_2^2 \\
        &\leq -\eta_t\nabla\ell(\Delta_{t})^T g(\Delta_{t},x_t) \\
        & \quad + \frac{1}{2}\eta_t^2L||g(\Delta_{t},x_t)||_2^2.
    \end{aligned}
    \label{inequa3}
\end{equation}
This inequality \ref{inequa3} shows that the expected decrease in the objective function yielded by the t-th step is bounded above by two components. One is the expected directional derivative of $\ell$ at $\Delta_{t}$ along $-g(\Delta_{t},x_t)$. The other is the second moment of $g(\Delta_{t},x_t)$.

\subsection{Proof of Total Differential }
\label{app:diff_details}
Given a function \( f(x_1, x_2, \dots, x_n) \) with variables \( x_1, x_2, \dots, x_n \), its total differential \( df \) can be expressed as:
\begin{equation}
df = \sum_{i=1}^{n} \frac{\partial f}{\partial x_i} dx_i, |dx_i|<\epsilon,
\end{equation}
where $\epsilon$ is a small value.
By considering the sample as a constant and treating $w$ as the input $x$, with $f = \ell(w)$ representing the loss calculation,  we can derive the total differential of $\ell(w)$ as follows:
\begin{equation}
    \partial\ell(w) = \sum_{i=n_1}^{n_2}\left\langle\frac{\partial\ell}{\partial w_i}, dw_{i}\right\rangle.
\label{eq:lossgap}
\end{equation}
$dw_{i}$ can be considered as the $\delta_{i}$ between $w_{i}$ and $\hat{w}_{i}$.
Therefore, we have:
\begin{equation}
\begin{aligned}
    \ell(\hat{w})-\ell(w)=\sum_{i=n_1}^{n_2}\left\langle\frac{\partial\ell}{\partial w_i},\delta_{i}\right\rangle,
\end{aligned}
\label{eq:6}
\end{equation}
$\left\langle\cdot,\cdot\right\rangle$ denotes inner product. 

\subsection{Efficient Implement of GAST}
In our method, one potential challenge lies in implementing data-layer selective gradient updating. We propose two approaches for implementing our algorithm: one that is \textit{computation-efficient} and the other that is \textit{memory-efficient}.

For the computation-efficient implementation, our goal is to minimize the amount of computation as much as possible. Specifically, for each Linear in every transformer layer during the backward pass, we save the intermediate per-sample's gradients. After sampling the training data points for that layer, we select and sum the saved mini-batch per-sample gradients. This approach introduces almost no additional computation, allowing the algorithm to be completed in a single forward-backward pass, thus maintaining computational efficiency. However, since it requires saving the per-sample gradients for all Linears in at least one transformer layer, it comes at the cost of increased memory consumption.

For the memory-efficient implementation, given that the computed gradient alignment score is a vector and only the sum of the scores for each Linear in every transformer layer is needed, we can reduce memory demands by releasing memory immediately after computing the score. An additional forward-backward pass can then be used to maintain the same memory consumption as vanilla PEFT training. Specifically, in the first forward-backward pass, we compute the gradient alignment score for each Linear, and then, after sampling data points for each layer, a second forward-backward pass is performed to select and merge the gradients.

These two implementation strategies allow our algorithm to adapt to various scenarios. Additionally, engineering techniques such as CPU offloading for per-sample gradients and asynchronous gradient aggregation have the potential to further improve the efficiency of the algorithm. In this paper, our focus is on the effectiveness of the algorithm itself rather than its implementation.

\section{Additional Details and Experiments}
\subsection{Dataset Statistics}
Detailed dataset statistics can be referred to Tab.~\ref{tab:dataset_description}. Note that we trained on Commonsense and Math10K for commonsense reasoning and arithmetic reasoning, respectively. During testing, we evaluated the predefined test sets of each dataset.Meanwhile, we showcase the instructions formats of different datasets in Tab.~\ref{tab:instruction_format_1} and Tab.~\ref{tab:instruction_format_2}.

\begin{table}[t]
\centering
\vspace{-5pt}
\setlength{\tabcolsep}{3pt}
\begin{tabular}
{lccr}
\toprule
Dataset  & \# Train & \# Test &Answer  \\\midrule
{\textbf{Commonsense}}        & 170K & -    &- \\
\quad{BoolQ}          & 9.4K &3,270    &Yes/No \\
\quad{PIQA}          & 16.1K&1,830    &Option \\
\quad{SIQA}         & 33.4K&1,954    &Option \\
\quad{HellaSwag}    & 39.9K&10,042   &Option \\
\quad{WinoGrande}     & 63.2K&1,267    &Option \\
\quad{ARC-e}          & 1.1K &2,376    &Option \\
\quad{ARC-c}         & 2.3K &1,172    &Option \\
\quad{OBQA}           & 5.0K &500      &Option \\\midrule
{\textbf{Math10K}}      & 10K & -    &- \\
\quad{GSM8K}        & 8.8K &1,319    &Number \\
\quad{AQuA}         & 100K &254      &Option \\
\quad{MAWPS}      & -    &238      &Number \\
\quad{SVAMP}         & -    &1,000    &Number \\
\bottomrule
\end{tabular}
\vspace{-2pt}
\caption{The statistics of datasets for evaluation. \#~Train and \#~Test denote the number of training and test samples respectively.}
\label{tab:dataset_description}
\end{table}

\subsection{Effect of Support Data Used in GAST}
\begin{figure}[t]
\centering
\begin{subfigure}[b]{0.49\linewidth}
    \centering
    \includegraphics[width=\linewidth]{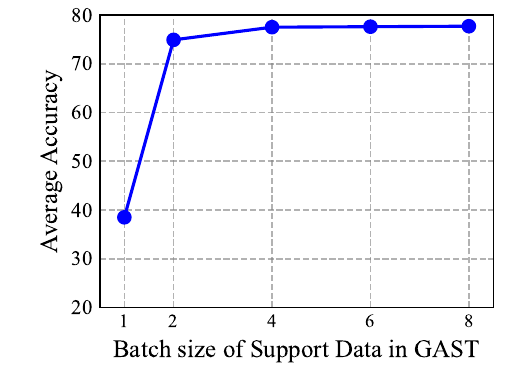}
    \captionsetup{aboveskip=0pt, belowskip=0pt}
\end{subfigure}%
\begin{subfigure}[b]{0.49\linewidth}
    \centering
    \includegraphics[width=\linewidth]{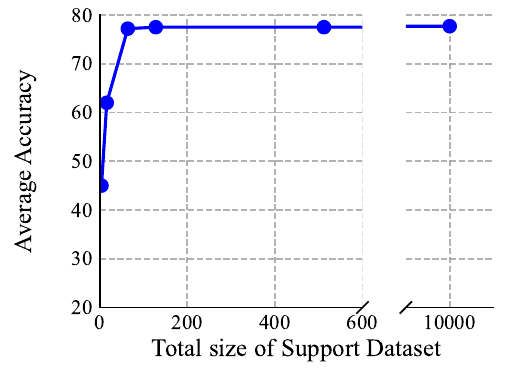}
    \captionsetup{aboveskip=0pt, belowskip=0pt}
\end{subfigure}
\caption{Additional experiment on support set.}
\label{fig:sup} 
\end{figure}

Fig~\ref{fig:sup} (left) illustrates the effect of varying batch size (i.e., $K$ in GAST) of support-set data on training a LLaMA 7B model using the commonsense dataset. We evaluated among five values: [1,2,4,6,8]. Experimental results show that the larger the value of $K$, the better the performance. However, after $K$ exceeds 4, the improvement becomes marginal. Considering that a larger $K$ also leads to increased computational overhead, we choose $K=4$ as the default parameter to balance computation and efficiency. Fig.~\ref{fig:sup} (right) shows the impact of different support set sizes on GAST, ranging from 16 to 10,000. For a fair comparison, we gradually increase the size of the support set used in GAST, but ensure that the training set and the validation set used for evaluation remain unchanged.
When the support set size exceeds 100, the benefit of further increasing the set size diminishes significantly.
This suggests that a small support set is sufficient to guide relatively accurate gradients. For simplicity and to eliminate the randomness of subset selection, we use the entire training set as the support set by default.
When training data contains noise or label errors, GAST may initially be affected if the support set is unrepresentative, as the gradient alignment mechanism could mistakenly treat relevant task features as noise. However, in our experiments in Section B.2, we specifically evaluated using the entire training set as the support set and observed strong robustness in such scenarios. This configuration helps ensure that the gradient alignment process remains stable and does not over-amplify the impact of noisy samples.

\begin{table}[]
\small
\centering
\setlength{\tabcolsep}{0.4mm}
\begin{tabular}{ccc}
\toprule
Method & Memory Cost & Training Time  \\\midrule
LoRA & 43.4 GB & 10 h\\
AdaLoRA & 46.7 GB & 13 h\\
IST & 36.6 GB  & 10 h\\
GREATS & 58.1 GB& 21 h\\
GAST (memory-efficient) & 51.5 GB& 19 h\\
GAST (computation-efficient) & 66.9 GB& 11.5 h\\\bottomrule
\end{tabular}
\caption{Computational cost of GAST on llama-7B with weight memory of 14GB and batch size of 16.}
\label{tab:complexity}
\end{table}
\subsection{Complexity Analysis}

\begin{table*}[t]
\centering
\caption{Performance comparison on common-sense reasoning benchmarks. Methods marked with * report results from the original papers.}
\label{tab:full_results_tab4}
\resizebox{\textwidth}{!}{%
\begin{tabular}{lccccccccc}
\toprule
Method & BoolQ & PIQA & SIQA & HellaSwag & WinoGrande & ARC-E & ARC-C & OBQA & AVG \\
\midrule
LoRA & 68.9 & 80.7 & 77.4 & 78.1 & 78.8 & 77.8 & 61.3 & 74.8 & 74.7 \\
LISA* & -- & -- & -- & -- & -- & -- & -- & -- & 75.3 \\
AdaLoRA* & -- & -- & -- & -- & -- & -- & -- & -- & 76.2 \\
RST & 68.3 & \textbf{82.3} & 78.1 & 76.6 & 80.1 & 79.9 & 63.3 & 77.8 & 75.8 \\
IST & 68.7 & 81.7 & 77.3 & 82.7 & 78.7 & 80.6 & 62.4 & 80.0 & 76.5 \\
GREATS & \textbf{69.6} & 81.1 & 78.0 & 81.0 & 79.8 & 79.5 & 63.8 & 77.6 & 76.3 \\
GAST (Random) & 68.0 & 81.4 & 79.1 & \textbf{84.8} & 79.9 & 79.7 & 63.1 & 79.2 & 76.9 \\
GAST (Top-K) & 50.4 & 63.7 & 79.3 & 77.0 & 37.7 & 79.3 & 63.8 & \textbf{80.4} & 66.4 \\
\midrule
GAST & 68.2 & 81.6 & \textbf{79.4} & 83.6 & \textbf{82.2} & \textbf{80.4} & \textbf{64.7} & 79.8 & \textbf{77.5} \\
\bottomrule
\end{tabular}%
}
\end{table*}
Tab~\ref{tab:complexity} presents the computational costs of different methods for training on Llama-7B under the same setting (weight memory 14GB, batch size 16, on 80G A100 GPU). Among all methods, LoRA exhibits the lowest memory cost (43.4 GB) and the fastest training time (7 hours), making it the most lightweight option. Data-selective method GREATS, on the other hand, requires significantly more memory (58.1 GB) and has the longest training time (21 hours), suggesting a higher computational overhead. GAST provides two variants to balance efficiency and resource use: the memory-efficient version uses less memory (51.5 GB) but at the cost of longer training (19 hours), while the computation-efficient version reduces training time to 11.5 hours, albeit with a higher memory footprint (66.9 GB). These results demonstrate that GAST enables flexible trade-offs between memory and computation.

\subsection{Full results of each subtask in the ablation study}
We provide the full results of all the subtasks in the ablation study in Tab~\ref{tab:full_results_tab4}. * indicate the results obtained from IST.

\begin{table*}[htbp]
\small
\centering
\begin{tabular}{p{15cm}l}
\toprule
\midrule
\rowcolor{gray!20} \textit{BoolQ} \\
\midrule
Please answer the following question with true or false, question: is house tax and property tax are same?\\ Answer format: true/false \\
\textit{the correct answer is true} \\
\midrule
\midrule
\rowcolor{gray!20} \textit{PIQA} \\
\midrule
Please choose the correct solution to the question: Extend life of flowers in vase. \\ Solution1: Add small amount of coffee in vase. \\ Solution2: Add small amount of 7UP in vase. \\ Answer format: solution1/solution2 \\
\textit{the correct answer is solution2} \\
\midrule

\midrule
\rowcolor{gray!20} \textit{SIQA} \\
\midrule
Please choose the correct answer to the question: Sydney walked past a homeless woman asking for change but did not have any money they could give to her. Sydney felt bad afterwards. How would you describe Sydney?\\Answer1: sympathetic Answer2: like a person who was unable to help Answer3: incredulous\\Answer format: answer1/answer2/answer3\\
\textit{the correct answer is answer1} \\
\midrule
\midrule
\rowcolor{gray!20} \textit{HellaSwag} \\
\midrule
Please choose the correct ending to complete the given sentence: Clean and jerk: A lady walks to a barbell. She bends down and grabs the pole. the lady\\Ending1: swings and lands in her arms. Ending2: pulls the barbell forward. Ending3: pulls a rope attached to the barbell. Ending4: stands and lifts the weight over her head.\\Answer format: ending1/ending2/ending3/ending4\\
\textit{the correct answer is ending4} \\
\midrule

\midrule
\rowcolor{gray!20} \textit{WinoGrande} \\
\midrule
Please choose the correct answer to fill in the blank to complete the given sentence: They were worried the wine would ruin the bed and the blanket, but the \_ was't ruined.\\Option1: blanket Option2: bed\\Answer format: option1/option2
\\
\textit{the correct answer is option2} \\
\midrule
\midrule
\rowcolor{gray!20} \textit{ARC-e} \\
\midrule
Please choose the correct answer to the question: Which piece of safety equipment is used to keep mold spores from entering the respiratory system?\\Answer1: safety goggles Answer2: breathing mask Answer3: rubber gloves Answer4: lead apron\\Answer format: answer1/answer2/answer3/answer4
\\
\textit{the correct answer is answer2} \\
\midrule
\midrule
\rowcolor{gray!20} \textit{ARC-c} \\
\midrule
Please choose the correct answer to the question: The end result in the process of photosynthesis is the production of sugar and oxygen. Which step signals the beginning of photosynthesis?\\Answer1: Chemical energy is absorbed through the roots. Answer2: Light energy is converted to chemical energy. Answer3: Chlorophyll in the leaf captures light energy. Answer4: Sunlight is converted into chlorophyll.\\Answer format: answer1/answer2/answer3/answer4
\\
\textit{the correct answer is answer4} \\
\midrule
\midrule
\rowcolor{gray!20} \textit{OBQA} \\
\midrule
Please choose the correct answer to the question: Predators eat\\Answer1: lions Answer2: humans Answer3: bunnies Answer4: grass\\Answer format: answer1/answer2/answer3/answer4
\\
\textit{the correct answer is answer3} \\
\midrule
\bottomrule
\end{tabular}
\small
\caption{Instructions format of commonsense reasoning task.}
\label{tab:instruction_format_1}
\end{table*}

\begin{table*}[htbp]
\small
\centering
\begin{tabular}{p{15cm}l}
\toprule
\midrule
\rowcolor{gray!20} \textit{GSM8K} \\
\midrule
Q: The farm has 30 cows and the zoo has 20 sheep.  The zoo has twice as many cows as the farm does.  The farm has half as many sheep as the zoo does.  How many animals do the farm and zoo have combined?\\
A: \textit{The farm has 30 cows and the zoo has 20 sheep. The zoo has twice as many cows as the farm, so the zoo has 30 cows. The farm has half as many sheep as the zoo, so the farm has 10 sheep. In total, the farm and zoo have 30 cows + 20 sheep + 10 sheep = 60 animals. The answer is 60.} \\
\midrule
Q: Rayna has 60 more dollars than Kassidy, and Kassidy has 3/4 times as much money as Aurelia. If Arelia has \$120, how much would each have if they decided to add together their money and share equally among them?\\
A: \textit{Rayna has 60 more dollars than Kassidy. That means Kassidy has 60 less than Rayna. Kassidy has 3/4 times as much money as Aurelia. That means Aurelia has 4/3 times as much money as Kassidy. The answer is 120.} \\
\midrule
\midrule
\rowcolor{gray!20} \textit{AQuA} \\
\midrule
Q: A rectangular piece of 150 sq m has a length which is 1m more than the 4 times the breadth. What is the perimeter of the piece? Answer Choices: (A) 60 m (B) 61 m (C) 62 m (D) 63 m (E) 64 m\\
A:  \textit{Let the breadth of the rectangular piece be x. Then the length of the piece would be 4x + 1. The area of the piece is 150 sq m, so 4x(4x + 1) = 150. This simplifies to 16x2 + 4x - 150 = 0. Solving this equation, we get x = 5. So the length of the piece is 4x + 1 = 21. The perimeter of the piece is 2(4x + 1) + 2x = 2(21) + 10 = 62 m.The answer is (C).}\\
\midrule
Q: Last year, 34 percent of Ace Book Company's sales revenue came from the sale of novels. Of the remaining revenue, 1/3 was from the sale of biographies. The company's revenue from the sale of novels was approximately, how many times its revenue from the sale of biographies? Answer Choices: (A) 1.3 (B) 1.5 (C) 2.1 (D) 2.5 (E) 3.1\\
A:  \textit{34\% of the total revenue was from the sale of novels. So the remaining 66\% was from the sale of biographies and other books. 66\% is equal to 2/3. So the revenue from the sale of novels was approximately 3 times the revenue from the sale of biographies.The answer is (E).}\\
\midrule
\midrule
\rowcolor{gray!20} \textit{MAWPS} \\
\midrule
Q: Zach wants to ride the Ferris wheel , the roller coaster , and the log ride . The Ferris wheel costs 2 tickets , the roller coaster costs 7 tickets and the log ride costs 1 ticket . Zach has 1 ticket . How many more tickets should Zach buy ?\\
A:  \textit{9.0}\\
\midrule
Q: In Shannon 's apartment complex , 0.16666666666666666 of the apartments are one - bedroom apartments and 0.3333333333333333 are two - bedroom apartments . What fraction of the apartments are either one- or two - bedroom apartments ?\\
A:  \textit{0.5}\\
\midrule
\midrule
\rowcolor{gray!20} \textit{SVAMP} \\
\midrule
Q: Jack received 9 emails in the morning, 10 emails in the afternoon and 7 emails in the evening. How many more emails did Jack receive in the morning than in the evening?\\
A: \textit{Jack received 9 emails in the morning, 10 emails in the afternoon and 7 emails in the evening. The difference between the number of emails he received in the morning and the number of emails he received in the evening is 9 - 7 = 2. The answer is 2.} \\
\midrule
Q: A mailman has to give 38 pieces of junk mail to each of the 78 blocks. If there are 19 houses on a block. How many pieces of junk mail should he give each house?\\
A: \textit{The mailman has to give 38 pieces of junk mail to each of the 78 blocks. That means he has to give 78 * 38 = 2964 pieces of junk mail in total. There are 19 houses on a block. So, each house should get 2964 / 19 = 156 pieces of junk mail. The answer is 156.} \\
\midrule
\bottomrule
\end{tabular}
\caption{Instructions format of arithmetic reasoning task.}
\label{tab:instruction_format_2}
\end{table*}

\end{document}